\documentclass{article}

\usepackage[utf8]{inputenc}
\usepackage[T1]{fontenc}

\usepackage{PRIMEarxiv}

\usepackage{amsmath,amssymb}
\usepackage{booktabs}
\usepackage{multirow}
\usepackage{graphicx}
\usepackage{xcolor}
\usepackage[section]{placeins}  

\usepackage[round]{natbib}

\usepackage{url}
\usepackage[hidelinks]{hyperref}

\title{CrowdCue: Specialist-Cue Conditioning for Vision-Language Crowd Counting%
\thanks{Preprint. Corresponding author: Moshiur Farazi (\texttt{moshiur.farazi@udst.edu.qa}).}}

\author{
  Moshiur Farazi\textsuperscript{1}, \enspace
  Bekir Ciftler\textsuperscript{1}, \enspace
  Abdulhalim Dandoush\textsuperscript{1}, \enspace
  Reda Bendraou\textsuperscript{1} \\[6pt]
  {\normalfont\normalsize \textsuperscript{1}University of Doha for Science and Technology, Doha, Qatar}
}

\date{}

\begin{document}
\maketitle

\begin{abstract}
Generative vision-language models (VLMs) offer a counting paradigm in which one model produces both a count and a natural-language account of the scene, yet their raw counting accuracy sits in the range of sub-million-parameter specialist regressors. The open question is whether auxiliary guidance from a pretrained specialist can lift them into useful territory, and through which channel that guidance is best routed. We evaluate Qwen2.5-VL-7B on four widely used crowd counting benchmarks (ShanghaiTech~A and~B, UCF-QNRF, NWPU-Crowd). Zero-shot prompting rarely produces a parseable count, so LoRA supervised fine-tuning establishes the baseline at overall MAE 81.64. Conditioning on a P2PNet-derived density heatmap as an auxiliary visual signal fails in every encoding we tested, and an adversarial-swap protocol shows the model reads the heatmap but applies it counterproductively. We propose \textit{CrowdCue}, a family that supplies the same specialist's already-integrated integer count to the VLM as a discrete symbol. The text-channel variant reaches MAE 72.04. The visual-channel variant, which renders the integer as printed digits and supplies it as a second image, reaches MAE \textbf{62.65}, the strongest result in this paper and well ahead of the cue-supplying specialist alone (84.45 on the same split). In the late-fusion VLM we study, the binding constraint is not the channel but the abstraction level at which the specialist signal is delivered.
\end{abstract}


\section{Introduction}\label{sec:intro}

A recurring question in computer vision is whether the generalist models reshaping the field have earned their place against the specialists they aim to replace. Crowd counting offers a sharp test case. The task is well-defined: estimate the number of people in an image. The specialist approach, density regression via convolutional and Transformer networks, has matured over a decade of work \citep{li2018csrnet, han2023steerer, ranasinghe2024crowddiff, ma2025zip} and now sits near MAE 48 on ShanghaiTech-A. Notably, the smallest variant of ZIP \citep{ma2025zip}, at 0.81M parameters, already reaches MAE 71.1, the same range a 7B-parameter VLM is likely to occupy under any near-term fine-tuning recipe. If raw MAE-per-parameter were the only criterion, the discussion of VLM-based counting would be short.

The reason it is not short is that generative VLMs such as Qwen2.5-VL \citep{bai2025qwen25vl} offer something density regressors do not: a unified model that produces a count, a structured reasoning trace, and a natural-language description in one forward pass. A reliable VLM counter would support deployments where the count is one of several requested outputs (e.g., crowd descriptions for accessibility, query-driven scene summaries, or operator-facing dashboards that justify their numbers). The narrow technical question this paper addresses is whether auxiliary spatial guidance, readily available from any pretrained specialist regressor, can lift the VLM's counting accuracy, and through which channel that guidance is best routed.

CrowdVLM-R1 \citep{wang2025crowdvlm} is the closest prior work, fine-tuning Qwen2.5-VL through SFT followed by GRPO with a fuzzy count reward. Their evaluation uses non-canonical datasets (i.e., sheep, GTAV characters, wheat heads, vehicles, street pedestrians), single-image input, and no auxiliary specialist signal. Like CrowdCLIP \citep{liang2023crowdclip}, CLIP-Count \citep{jiang2023clipcount}, and ProgRoCC \citep{jiang2025progrocc}, the input pathway is one image and a text prompt; the auxiliary-conditioning question has not been studied for generative VLM counting.

Our initial hypothesis followed the multi-modal crowd counting literature: provide the VLM with a density heatmap as a second image. Heatmaps are cheap to generate, explicitly encode where people concentrate, and have been shown to help when fused with broker representations in dedicated counting architectures \citep{meng2024broker, meng2025freelunch}. Three principal encodings on valid P2PNet-derived heatmaps (dual-image, side-by-side concatenation, and alpha-blended overlay) all degrade MAE relative to the single-image SFT model. The adversarial-swap protocol on the dual-image checkpoint produces a 23.4-point per-condition MAE spread, and the worst condition is a content-coherent but spatially mismatched (shuffled) heatmap at 119.65 (Section~\ref{sec:adversarial-results}). The model therefore reads the secondary image but applies its spatial structure incorrectly. The pattern is in the same family as the multi-image underuse documented by MuirBench \citep{wang2024muirbench} across the VLM family, though qualitatively distinct. Where MuirBench characterises under-attention to secondary images, our finding is active misuse of attended-to spatial signal. We attribute both to the late-fusion architecture in which each image is encoded independently through the ViT, with cross-image attention deferred to the LLM decoder where the model has not learned to decode a dense spatial field into a count.

The diagnosis suggests the resolution. If the auxiliary signal cannot be decoded as a dense field, the next attempt should deliver it in a pre-integrated form: an integer estimate produced by the same specialist. We instantiate this as the \textit{CrowdCue} family. CrowdCue-T inserts the integer into the prompt text as a Chameleon-style \citep{lu2023chameleon} specialist composition and reaches MAE 72.04. To test whether the text channel is itself the explanation, we then train CrowdCue-V, which renders the same integer as printed digits and supplies it through the visual channel as a second image. If the channel hypothesis is right, CrowdCue-V should fail in the regime of the dense-heatmap encodings. It does not: CrowdCue-V reaches MAE \textbf{62.65}, the lowest result in the paper, with an aggregate margin of 9.4 MAE over CrowdCue-T that is concentrated in the extreme-density tail. The visual channel itself is therefore not the binding constraint; signal abstraction is. Adversarial overrides on both variants confirm causal use, with MAE spreads of 304 (CrowdCue-T) and 877 (CrowdCue-V), both more than an order of magnitude larger than the 23.4-point spread under valid heatmap swap.

The paper's contributions are best read in this order:

\begin{itemize}
\item To our knowledge, we present the first fine-tuned generative-VLM evaluation spanning four widely used crowd counting benchmarks (ShanghaiTech A/B \citep{zhang2016shanghaitech}, UCF-QNRF \citep{idrees2018ucfqnrf}, NWPU-Crowd \citep{wang2020nwpu}; 6{,}342 images, counts 0--20{,}033). LoRA SFT lifts format compliance from 0.8--13.4\% to 100\% and yields MAE 81.64 overall (70.17 on SHA-A), placing the bare model in the accuracy regime of sub-million-parameter specialist regressors.
\item We document the failure of dense-density-heatmap conditioning across three principal visual encodings on a verified-valid P2PNet-derived signal, and confirm it with an adversarial swap protocol (23.4-point MAE spread, shuffled heatmap worst at 119.65). The protocol extends naturally to other VLM multi-image conditioning questions and doubles as a signal-validity check.
\item We propose the \textit{CrowdCue} family: specialist-cue conditioning in which a pretrained specialist's integer count is supplied to the VLM as an already-integrated symbol. CrowdCue-T routes the integer through the text channel and reaches MAE 72.04; CrowdCue-V renders the same integer as a printed digit image and reaches MAE \textbf{62.65}, the lowest result in the paper (Sections~\ref{sec:textinject-results}, \ref{sec:abstraction-controls}). Together the two variants provide evidence that signal abstraction, not channel modality, is what limits auxiliary-signal usability in our late-fusion setting.
\item We characterise where VLM counting works and where it does not: sparse-scene MAE 8.99, zero-count accuracy 94.3\% (a failure mode density regressors are prone to), and a superlinear error curve above 1{,}000 people that our model does not resolve.
\end{itemize}

\section{Related Work}\label{sec:related}

Table~\ref{tab:litreview} summarises the methods discussed in this section, grouped into three families separated by horizontal rules: specialist crowd-counting architectures, language-vision counters, and multi-image vision language models (VLMs) diagnostics with their tool-use counterparts.

\begin{table}[!htbp]
\caption{Summary of related work. The three families are separated by horizontal rules.}\label{tab:litreview}
\centering
\small
\begin{tabular}{@{}l l l p{0.40\columnwidth}@{}}
\toprule
Method & Year & Paradigm & Key contribution \\
\midrule
ZIP \citep{ma2025zip} & 2025 & Density & Zero-Inflated Poisson loss; SHA-A 47.8 (105M), 71.1 (0.81M) \\
DSGC-Net \citep{wu2025dsgcnet} & 2025 & Density & Dual-stream graph CNN; SHA-A 48.9, SHB 5.9 \\
RCCFormer \citep{chen2025rccformer} & 2025 & Density & Multi-feature Transformer; SHA 48.3, QNRF 77.6, NWPU 74.3 \\
TCFormer \citep{liu2025tcformer} & 2025 & Density & 5M-parameter weakly-supervised Transformer \\
CrowdDiff \citep{ranasinghe2024crowddiff} & 2024 & Diffusion & Multi-hypothesis density; SHA-A 47.4 \\
STEERER \citep{han2023steerer} & 2023 & Density & Selective inheritance for scale variation \\
P2PNet \citep{song2021p2pnet} & 2021 & Detection & Point detection; our auxiliary specialist (Sec.~\ref{sec:textinject}) \\
CSRNet \citep{li2018csrnet} & 2018 & Density & Dilated convolutions on VGG-16; foundational \\
MCNN \citep{zhang2016shanghaitech} & 2016 & Density & Multi-column CNN; introduced ShanghaiTech \\
\midrule
CrowdCLIP \citep{liang2023crowdclip} & 2023 & CLIP & Unsupervised counting via text-image ranking \\
CLIP-Count \citep{jiang2023clipcount} & 2023 & CLIP & Zero-shot text-guided density maps \\
CLIP-EBC \citep{clipEBC2025} & 2025 & CLIP & Blockwise classification; SHA-A 55.0 \\
ProgRoCC \citep{jiang2025progrocc} & 2025 & CLIP & Progressive coarse-to-fine; rough annotations \\
CrowdVLM-R1 \citep{wang2025crowdvlm} & 2025 & Gen.\ VLM & SFT + GRPO fuzzy reward; single-image; non-canonical data \\
\textbf{Ours} & 2026 & Gen.\ VLM & Fine-tuned VLM eval on four benchmarks; specialist-cue conditioning (text + visual) \\
\midrule
CAPTURE \citep{pothiraj2025capture} & 2025 & Diagnostic & VLM counting under occlusion fails; auxiliary info helps \\
MuirBench \citep{wang2024muirbench} & 2024 & Diagnostic & Multi-image underuse across VLMs; GPT-4o 68\%, open $<$33\% \\
Chameleon \citep{lu2023chameleon} & 2023 & Tool-use & LLM planner composing specialists via the text channel \\
Visual Sketchpad \citep{hu2024sketchpad} & 2024 & Visual tool-use & Specialist outputs as visual artifacts; image-channel contrast \\
VTool-R1 \citep{vtoolr1_2026} & 2026 & Visual CoT & VLMs learn visual tool use via RL \\
\bottomrule
\end{tabular}
\end{table}

\subsection{Crowd Counting: From Specialist to Generalist}

Density regression remains the dominant paradigm. CSRNet \citep{li2018csrnet} established the template (a VGG-16 frontend feeding dilated convolutions over an integrated density map), and a decade of work since has refined the recipe through context-aware fusion \citep{liu2019context}, Bayesian point-loss formulations \citep{ma2019bayesian}, selective inheritance for scale variation \citep{han2023steerer}, and diffusion-based multi-hypothesis estimation \citep{ranasinghe2024crowddiff}. The 2025 frontier sits near SHA-A MAE 48: ZIP \citep{ma2025zip} reaches 47.8 by replacing MSE with a Zero-Inflated Poisson likelihood, DSGC-Net \citep{wu2025dsgcnet} 48.9 via dual-stream graph convolutions, and RCCFormer \citep{chen2025rccformer} 48.3 with a multi-feature Transformer. Notably, ZIP's lightweight ladder reaches MAE 71.1 with a 0.81M-parameter backbone, which we return to when situating our 7B-parameter VLM (Section~\ref{sec:context}). P2PNet \citep{song2021p2pnet} took a different route, replacing continuous density with discrete point regression for joint counting and localisation; we use a pretrained P2PNet as the auxiliary specialist whose count estimate we inject in Section~\ref{sec:textinject}.

A parallel line of work brings vision-language representations into counting. CrowdCLIP \citep{liang2023crowdclip} uses CLIP's discriminative embeddings for unsupervised counting via text-image matching. CLIP-Count \citep{jiang2023clipcount} generates zero-shot density maps guided by text prompts. ProgRoCC \citep{jiang2025progrocc} combines CLIP with progressive coarse-to-fine estimation. CLIP-EBC \citep{clipEBC2025} adapts CLIP for blockwise count classification, reaching SHA-A MAE 55.0 and SHB 6.3. These methods sit at the boundary between specialist and generalist: they use language-vision alignment but remain purpose-built counting models that do not generate free-form reasoning.

Closest to our work, CrowdVLM-R1 \citep{wang2025crowdvlm} fine-tunes Qwen2.5-VL through SFT followed by GRPO with a fuzzy count reward, evaluated on five non-canonical datasets (i.e., sheep, GTAV characters, wheat heads, vehicles, street pedestrians). Two differences are central. CrowdVLM-R1 uses single-image input throughout (no auxiliary spatial or numerical signal) and its counts cap below 3{,}500 per image, while NWPU-Crowd reaches 20{,}033. We extend the setting on both axes: widely used crowd counting benchmarks, and explicit auxiliary conditioning, with our central methodological move being the integer cue rather than a dense second image. Whether the fuzzy-reward GRPO formulation transfers cleanly to our 7B LoRA / canonical-benchmark setting is left to future work (Section~\ref{sec:conclusion}).

\subsection{Multi-Image Reasoning and Its Failure Modes}

Our heatmap conditioning experiments require the VLM to fuse spatial information across two distinct images. Recent diagnostic work suggests this is structurally difficult for current architectures. \citet{pothiraj2025capture} show that VLM counting accuracy degrades sharply even under simple pattern occlusion, for models as capable as GPT-4o and Qwen2-VL. MuirBench \citep{wang2024muirbench} extends the picture across 12 multi-image reasoning tasks and 11{,}264 images: GPT-4o reaches 68.0\% accuracy and Gemini Pro 49.3\%, both well below the human baseline of 93.1\%, while open-source single-image-trained VLMs score under 33\%. Critically, MuirBench documents a positional bias verified across multiple VLM families: images placed mid-prompt receive less effective attention than images embedded in answer options. A density heatmap presented as the second image in our prompt sits in exactly this attention-disadvantaged position. \citet{hayat2025attention} identify a related mechanism: VLMs discard visual evidence in favour of language priors when the text pathway alone suffices to minimise training loss.

On the visual prompting side, Set-of-Mark (SoM) \citep{yang2023setofmark} demonstrates that overlaying alphanumeric markers directly onto images improves VLM grounding for sparse object detection. Our alpha-blended overlay results (Section~\ref{sec:adversarial-results}) suggest that overlay-style conditioning does not transfer to dense crowd counting, though we test blended density fields rather than SoM-style discrete markers. VTool-R1 \citep{vtoolr1_2026} takes a different approach, training VLMs to generate their own visual reasoning steps through tool use under reinforcement learning. The takeaway across these efforts: auxiliary visual information is hard for current VLMs to use without explicit optimisation pressure or architectural change; our text-channel route bypasses this constraint entirely.

\subsection{Tool-Use and Text-Channel Specialist Composition}

The pattern of composing a specialist module with a general reasoning model through the text channel predates the multimodal era. Chameleon \citep{lu2023chameleon} formalises it: an LLM acts as a planner, sequencing calls to off-the-shelf vision models, web search, Python, and heuristic modules, then composing their textual outputs into a final answer. The architectural choice, returning the specialist's output as text rather than as a tensor or image, sidesteps cross-modal alignment entirely. Visual Sketchpad \citep{hu2024sketchpad} explores the alternative: it renders specialist outputs (i.e., detection boxes, segmentation masks) onto an image canvas that the VLM then reads. This image-channel framing recovers spatial information but inherits exactly the multi-image reasoning weakness MuirBench documents. We position the text-channel injection studied in Section~\ref{sec:textinject} as a Chameleon-style specialist composition, narrowed to the single signal most relevant to crowd counting: an integer count from a specialist regressor.

\section{Method}\label{sec:method}

\begin{figure}[htbp]
\includegraphics[width=1\columnwidth]{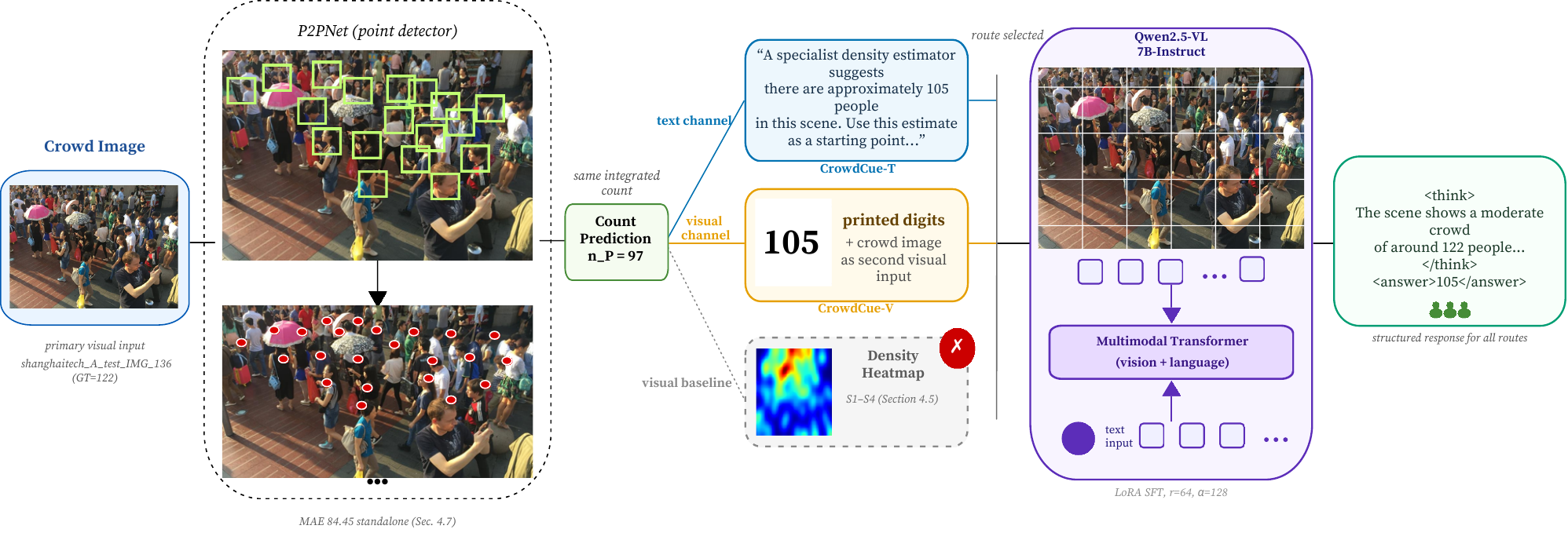}
\caption{Overview of the evaluation framework. A crowd image is optionally paired with a P2PNet-derived density heatmap and fed to Qwen2.5-VL-7B under one of two training regimes. We evaluate three heatmap conditioning strategies against the single-image baseline. The CrowdCue variants (Section~\ref{sec:textinject}) replace the heatmap with a specialist integer estimate.}
\label{fig:pipeline}
\end{figure}

We evaluate Qwen2.5-VL-7B-Instruct \citep{bai2025qwen25vl} for crowd counting under two training regimes (zero-shot, SFT), three heatmap conditioning strategies, and the two-variant CrowdCue family (Section~\ref{sec:textinject}), which supplies a specialist's integer estimate through the text or the visual channel. Fig.~\ref{fig:pipeline} illustrates the evaluation framework.

\subsection{Task Formulation}

Given a crowd image $I$, the model generates a structured response:
\begin{equation}
\texttt{<think>} \; r \; \texttt{</think>} \; \texttt{<answer>} \; \hat{y} \; \texttt{</answer>}
\end{equation}
where $r$ is a free-text reasoning trace and $\hat{y} \in \mathbb{Z}_{\geq 0}$ is the predicted count. The predicted count is extracted via the regular expression \texttt{<answer>(\textbackslash d+)</answer>}. If the model fails to produce a valid match, the output is marked as a format failure.

\subsection{Heatmap Generation}\label{sec:heatmap}

\begin{figure}[htbp]
\includegraphics[width=1\columnwidth]{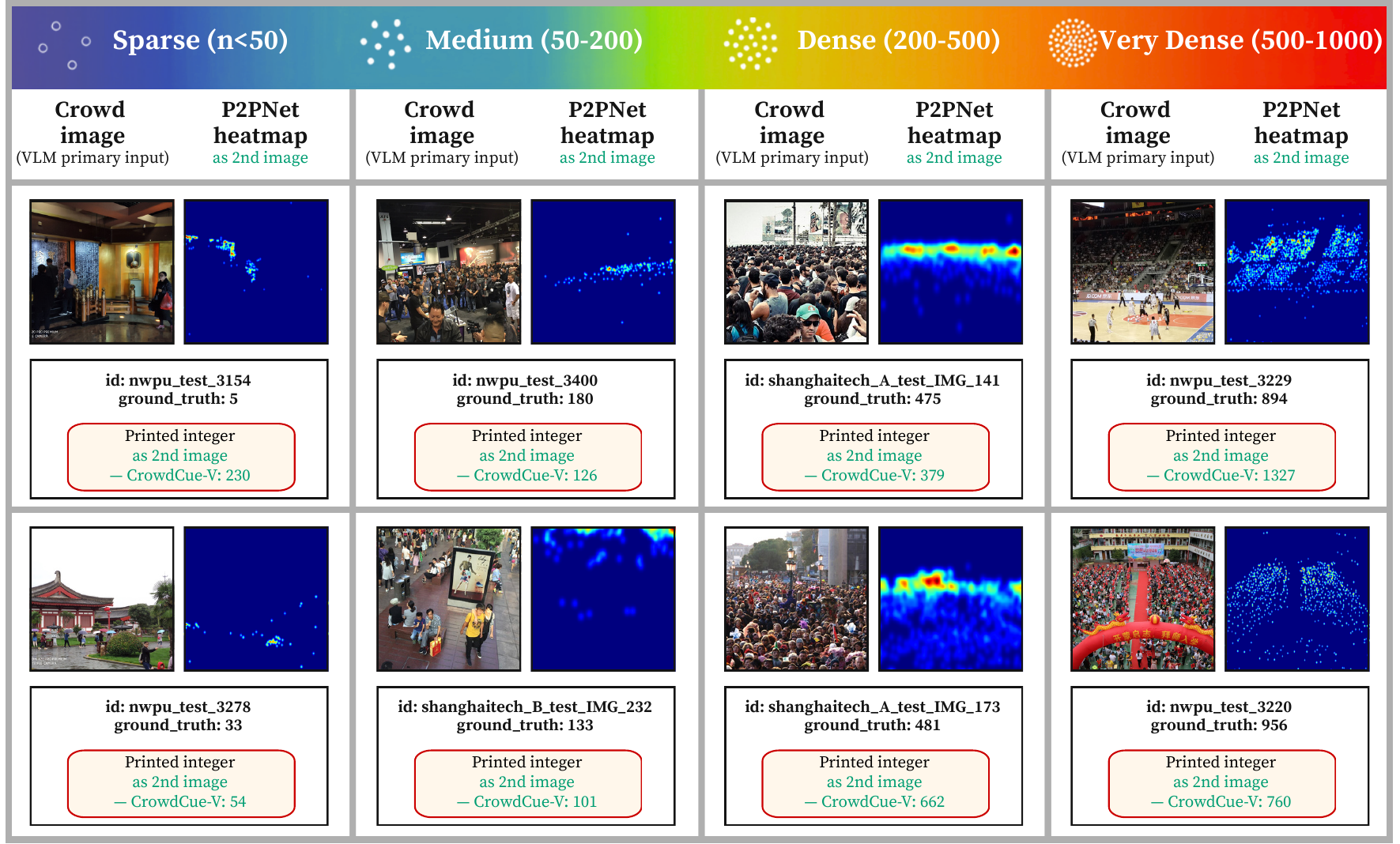}
\caption{Examples of P2PNet-derived density heatmaps across four density levels, with the corresponding printed-digit cue used by CrowdCue-V. Warm colours (red, yellow) indicate high crowd density; cool colours (blue, green) indicate low density.}
\label{fig:heatmaps}
\end{figure}

For each crowd image, we render a pseudo-density heatmap from the point detections of a pretrained P2PNet \citep{song2021p2pnet} (official ShanghaiTech-A checkpoint applied uniformly across the four datasets). The detected point set is converted to a continuous density field by Gaussian splatting at a fixed bandwidth of $\sigma = 15$ pixels, the fixed-kernel convention used for ShanghaiTech Part-B ground truth (Part-A ground truth conventionally uses geometry-adaptive kernels). The resulting density map is resized to $448 \times 448$, normalised to $[0,1]$ (a choice that preserves spatial structure but discards the absolute scale of the density field), and rendered with the \texttt{jet} colormap, producing an RGB image where red indicates high density and blue indicates low density. All 6{,}342 heatmaps are generated offline; Fig.~\ref{fig:heatmaps} shows examples across four density levels.

\subsection{Training Regime 1: Zero-Shot Prompting}

We evaluate the base Qwen2.5-VL-7B-Instruct model without any task-specific training. The model receives a crowd image (optionally with a heatmap) and a question asking it to count the people, with instructions to format the output as specified.

\subsection{Training Regime 2: Supervised Fine-Tuning}\label{sec:sft}

We fine-tune Qwen2.5-VL-7B-Instruct using Low-Rank Adaptation (LoRA) \citep{hu2022lora} with rank $r=64$, scaling factor $\alpha=128$, and dropout 0.05, applied to all attention and feed-forward projection matrices (\texttt{q, k, v, o, gate, up, down}). The target list matches modules in both the language model and the vision tower (196 and 96 projection matrices respectively), so the visual encoder is adapted rather than frozen. This yields 134M trainable parameters (1.8\% of the 7.6B total). Training uses the Adam optimiser with cosine learning rate schedule (peak $10^{-4}$, 5\% warmup), effective batch size 8 (per-device 1, gradient accumulation 8), bf16 precision, and gradient checkpointing on a single NVIDIA H100 80GB GPU. Each run trains for 3 epochs over 5,010 samples (1,881 steps), completing in approximately 3.7 hours.

Input images are resized to a maximum of $512 \times 512$ pixels, producing roughly 800 visual tokens per image. This constraint is imposed by the interaction between Qwen2.5-VL's dynamic tokenisation and the model's effective context length. Training data pairs each image with a question (randomly sampled from a template pool of 7--8 variants to prevent template memorisation) and a target response containing a density-appropriate reasoning trace and the ground-truth count.

\subsection{Density Heatmap Conditioning Strategies}\label{sec:strategies}

We evaluate three strategies for incorporating the heatmap as a visual signal:

\paragraph{S1: Dual-image input.} The crowd image and heatmap are provided as two separate images in the VLM prompt. A text preamble explains the heatmap's colour semantics. The ViT encoder processes each image independently; cross-image attention occurs only in the LLM decoder.

\paragraph{S2: Overlay blending ($\alpha = 0.4$).} The heatmap is alpha-blended onto the crowd image, producing a single composite image in which density colouring is spatially co-located with the crowd.

\paragraph{S3: Side-by-side concatenation.} The crowd image and heatmap are placed side by side in a single wide image. Both modalities pass through the ViT in one forward pass, but the crowd image is halved in resolution.

Each conditioning strategy is paired with an SFT training run using the same hyperparameters as the single-image baseline. For S1, we use the \texttt{dual} conversation template; for S2 and S3, we compose the image at loading time and use a single-image template with an appropriate text preamble.

\subsection{Adversarial Visual Ablation}\label{sec:adversarial}

To test whether the model causally uses the heatmap, we replace the correct heatmap in the dual-image model (S1) with adversarial inputs during evaluation:

\begin{itemize}
\item \textbf{Random noise}: Jet-coloured random pixels with no spatial structure.
\item \textbf{Shuffled}: A heatmap from a different, randomly selected image (correct colour distribution, wrong spatial alignment).
\item \textbf{Inverted}: The correct heatmap with hot$\leftrightarrow$cold colours swapped (correct spatial structure, opposite density encoding).
\item \textbf{Uniform gray}: A constant mid-gray image (zero information).
\end{itemize}

If the model uses the heatmap, replacing it with adversarial inputs should measurably change the predicted counts. If the model ignores it, all conditions should produce similar MAE.


\subsection{CrowdCue: Specialist-Cue Conditioning}\label{sec:textinject}

The diagnosis from Sections~\ref{sec:strategies} and~\ref{sec:adversarial} is that the late-fusion VLM does not effectively consume the secondary image when its content is a dense density field. The Chameleon framework \citep{lu2023chameleon} suggests a remedy that bypasses this bottleneck: deliver the specialist's output as a discrete already-integrated symbol rather than a pre-integration tensor. We instantiate this idea as a family of methods we call \textit{CrowdCue}. A pretrained P2PNet \citep{song2021p2pnet} is run over each crowd image and its retained head detections are summed into a single integer count $\hat{n}_{P}$; that integer is then supplied to the VLM as a discrete cue. The family admits two channel variants:

\paragraph{CrowdCue-T (text channel).} The integer is inserted into the question text as natural language:

\medskip
\noindent\textit{``A specialist density estimator suggests there are approximately $\hat{n}_{P}$ people in this scene. Use this estimate as a starting point and look at the image carefully; verify, adjust, or override the estimate as needed.''}
\medskip

The VLM receives the original crowd image as its only visual input. This is the variant we initially designed, and the one we use throughout Sections~\ref{sec:textinject-results} and~\ref{sec:ablations} unless we specify otherwise.

\paragraph{CrowdCue-V (visual channel).} The same integer $\hat{n}_{P}$ is rendered as printed digits on a plain white background ($448 \times 448$, monospaced font, jet-free) and supplied as a second image alongside the crowd photograph. The question text does not name the count. We introduce this variant in Section~\ref{sec:abstraction-controls} as a mechanism control: if the text channel is what works, CrowdCue-V should fail in the same regime as the density-heatmap encodings of Section~\ref{sec:strategies}.

We use the official P2PNet ShanghaiTech-A checkpoint for all four datasets, accepting the resulting distribution-shift error on SHB, QNRF, and NWPU; standalone accuracy of the injected signal is reported in Section~\ref{sec:textinject-results}. This makes the injected estimate an imperfect prior. The model has to learn when to trust it and when to override it, which is the core capability we want from the family.

During SFT the assistant target stays minimal: a brief scene descriptor in the \texttt{<think>} block (the same single-image bucket descriptor used for the SFT-single baseline) followed by the ground-truth count in the \texttt{<answer>} tag. The training target makes no explicit reference to the injected estimate. The model is free to discover, from the gradient signal alone, that the cue in the user prompt is highly correlated with the target answer and is worth anchoring on. We investigate an alternative target that explicitly models verify-and-adjust behaviour in Section~\ref{sec:ablations}; the simpler flat target reaches strictly better MAE.

To test whether the model uses the cue causally (mirroring the adversarial protocol of Section~\ref{sec:adversarial}), we evaluate four override conditions at inference time, replacing $\hat{n}_{P}$ with:

\begin{itemize}
\item \textbf{Zero}: $\hat{n}_{P} \leftarrow 0$. If the model trusts the cue blindly, predicted counts collapse to near zero.
\item \textbf{Random}: $\hat{n}_{P} \leftarrow \mathrm{Uniform}\{0,\ldots,1000\}$ (fixed seed). Tests whether the model treats noise as informative.
\item \textbf{Ten times correct}: $\hat{n}_{P} \leftarrow 10 \cdot \hat{n}_{P}$. Tests upper-bound sensitivity.
\item \textbf{Half correct}: $\hat{n}_{P} \leftarrow \lfloor \hat{n}_{P} / 2 \rfloor$. Tests fine-grained sensitivity.
\end{itemize}

A model that ignores the cue (or, as the dual-image model with valid heatmaps does, reads the auxiliary signal but cannot convert it into accuracy) will produce a similar MAE across all four overrides. A model that uses it causally will shift its predicted distribution in the direction of the injected number, with the override-to-correct MAE gap quantifying the strength of that anchoring effect. We apply the full four-override protocol to both variants of the family.

\section{Experiments}\label{sec:experiments}

\subsection{Datasets}

We evaluate on four widely used crowd counting benchmarks (Table~\ref{tab:datasets}). NWPU-Crowd test labels are not public, so we use its 500-image validation split as the test set throughout; every external comparison against published NWPU test-server numbers is flagged where it occurs.

\begin{table}[!htbp]
\caption{Dataset summary. NWPU-Crowd test labels are not public; we use the validation split as test.}\label{tab:datasets}
\centering
\begin{tabular}{@{}lrrrl@{}}
\toprule
Dataset & Train & Test & Count Range & Scene Type \\
\midrule
ShanghaiTech A & 300 & 182 & 33--3{,}138 & Dense gatherings \\
ShanghaiTech B & 400 & 316 & 9--576 & Street scenes \\
UCF-QNRF & 1{,}201 & 334 & 49--12{,}865 & Web-crawled crowds \\
NWPU-Crowd & 3{,}109 & 500 & 0--20{,}033 & Diverse scenes \\
\midrule
\textbf{Total} & \textbf{5{,}010} & \textbf{1{,}332} & & \\
\bottomrule
\end{tabular}
\end{table}

Together, the training set spans five density levels: sparse (0--50 people, 16.3\%), medium (50--200, 31.7\%), dense (200--500, 28.4\%), very dense (500--1{,}000, 12.9\%), and extreme ($>$1{,}000, 10.9\%). NWPU-Crowd uniquely includes 35 test images with zero people, testing the model's ability to output a null count.

\subsection{Evaluation Protocol}

At inference time we use greedy decoding (\texttt{do\_sample=False}) with a maximum of 256 generated tokens. We parse the predicted count from the output using the regular expression \texttt{<answer>(\textbackslash d+)</answer>}. If parsing fails, the output is flagged as a format failure and the prediction is set to 0. The exception is the zero-shot setting, where compliance is so low that imputed zeros would swamp the metric; zero-shot MAE is computed over parseable outputs only and flagged as non-comparable (Section~\ref{sec:zeroshot}). We report Mean Absolute Error (MAE) per dataset and overall, with Root Mean Squared Error (RMSE) for ShanghaiTech-A (Table~\ref{tab:main}) and overall for each conditioning configuration (Tables~\ref{tab:visual-encodings-valid} and~\ref{tab:abstraction-controls}).

\subsection{Zero-Shot Evaluation: The Format Barrier}\label{sec:zeroshot}

Without fine-tuning, Qwen2.5-VL-7B-Instruct fails to produce a parseable \texttt{<answer>N</answer>} tag for the vast majority of test images. Table~\ref{tab:format} reports the format success rates.

\begin{table}[!htbp]
\caption{Format compliance: fraction of test images producing a valid parseable count.}\label{tab:format}
\centering
\begin{tabular}{lcc}
\toprule
Configuration & Valid Format & Rate \\
\midrule
Zero-shot, single image & 179 / 1{,}332 & 13.4\% \\
Zero-shot, dual image & 11 / 1{,}332 & 0.8\% \\
\midrule
SFT, single image & 1{,}332 / 1{,}332 & 100\% \\
SFT, dual image & 1{,}332 / 1{,}332 & 100\% \\
\bottomrule
\end{tabular}
\end{table}

The base model typically generates extended prose descriptions, embeds counts within conversational text, or refuses to commit to a specific number. Adding a heatmap as a second image worsens format compliance from 13.4\% to 0.8\%, suggesting the additional visual complexity further disrupts the model's ability to follow output formatting instructions. Since the zero-shot MAE is computed only over the small fraction of parseable outputs (179 or 11 images), these numbers are not comparable to the SFT results and we do not report them as valid baselines. SFT resolves the format problem completely, achieving 100\% compliance across all 1{,}332 test images.

\subsection{SFT Baseline Results}\label{sec:sft-results}

Training loss for dual-image SFT drops sharply in the first 5\% of training (16.5 $\to$ 6.5), indicating rapid format learning, then stabilises through three epochs (1{,}881 steps over 5{,}010 samples) with no sign of overfitting.

Table~\ref{tab:main} presents the SFT results for single-image and dual-image configurations.

\begin{table}[!htbp]
\caption{SFT results on four benchmarks. Best MAE per dataset in \textbf{bold}. The single-image baseline outperforms the dual-image model with a valid P2PNet-derived heatmap on three of four datasets and overall.}\label{tab:main}
\centering
\begin{tabular}{@{}lcccccc@{}}
\toprule
& \multicolumn{2}{c}{SHA-A} & SHA-B & QNRF & NWPU & Overall \\
\cmidrule(lr){2-3}
Config & MAE & RMSE & MAE & MAE & MAE & MAE \\
\midrule
SFT single & \textbf{70.17} & 116.01 & 15.47 & \textbf{119.19} & \textbf{102.55} & \textbf{81.64} \\
SFT dual & 71.16 & 114.45 & \textbf{13.93} & 153.61 & 119.13 & 96.26 \\
\bottomrule
\end{tabular}
\end{table}

The single-image SFT model achieves MAE 81.64 overall, with its strongest performance on ShanghaiTech~B (MAE 15.47), a dataset of moderate-density street scenes. On the high-density ShanghaiTech~A, the model reaches MAE~70.17, within the range of early density regression methods like CSRNet (68.2) though substantially behind recent methods such as CLIP-EBC (55.0) and CrowdDiff (47.4).

The dual-image model, which receives both the crowd image and the P2PNet-derived density heatmap as a second image, performs worse than single-image on three of four datasets, with an overall MAE of 96.26 (+18\% relative degradation). The sole exception is ShanghaiTech~B, where dual-image achieves a small improvement (13.93 vs.\ 15.47). We investigate the negative result in detail in Section~\ref{sec:adversarial-results}.

\subsection{Adversarial Ablation: The Model Attends but Cannot Exploit}\label{sec:adversarial-results}

We retrained the dual, side-by-side, and overlay configurations of Section~\ref{sec:strategies} with the canonical P2PNet heatmap (Section~\ref{sec:heatmap}). Two valid-signal results determine the interpretation of the visual-channel failure.

First, three independent encodings of the valid P2PNet-derived heatmap as a second-image conditioning signal all degrade overall MAE relative to the single-image baseline (Table~\ref{tab:visual-encodings-valid}): dual (heatmap as a separate tensor), side-by-side (heatmap concatenated horizontally with the crowd), and overlay (heatmap alpha-blended at $\alpha=0.4$). The overlay configuration collapses to MAE 267.45, driven entirely by the extreme-density bucket on NWPU (per-bucket MAE 2{,}361) where the model emits pathological repeating tokens; dual and side-by-side, which preserve the crowd image as a distinct tensor, degrade more gracefully but still fail the baseline by 15--23 MAE. Format compliance stays at 100\% in the dual and side-by-side runs and 1{,}331 of 1{,}332 in the overlay run, and the overlay MAE computed over parseable outputs alone is 266.86, so the collapse reflects decoding failure rather than a parsing artefact.

\begin{table}[!htbp]
\caption{Three visual encodings of the (valid) P2PNet density heatmap as SFT auxiliary signal. All degrade MAE relative to the no-heatmap baseline. Trained separately, evaluated on the combined 1{,}332-image test split. The final column is overall RMSE across all test images.}\label{tab:visual-encodings-valid}
\centering
\begin{tabular}{lcccccc}
\toprule
Configuration & SHA-A & SHB & QNRF & NWPU & Overall & RMSE \\
\midrule
SFT-single (no heatmap, baseline) & 70.17 & 15.47 & 119.19 & 102.55 & 81.64 & 352.66 \\
SFT-dual (valid P2PNet heatmap) & 71.16 & 13.93 & 153.61 & 119.13 & 96.26 & 443.13 \\
SFT-side-by-side (valid heatmap) & 75.80 & 16.84 & 156.40 & 136.63 & 104.86 & 497.93 \\
SFT-overlay $\alpha = 0.4$ (valid heatmap) & 73.37 & 17.66 & 170.30 & 560.86 & 267.45 & 4586.92 \\
\bottomrule
\end{tabular}
\end{table}

Second, the same five-condition adversarial swap protocol applied to the dual-image SFT now produces a spread of 23.39 MAE points across the five heatmap conditions (Table~\ref{tab:adversarial}), with the shuffled condition (a heatmap from a different randomly-selected test image, content-coherent but spatially mismatched) the worst at 119.65. Inverted (hot/cold swap with the same spatial structure) moves MAE by only 1.6 points; gray (uniform, no spatial information) by 5.5 points; noise by 6.9; shuffled by 23.4. The model reads spatial structure from the heatmap, but applies it incorrectly: presenting confidently-wrong spatial information harms more than presenting noise.

Two further probes bound the failure. Replacing the P2PNet-derived heatmap with one splatted from the ground-truth annotations at evaluation time improves the dual model only to 91.05, still 9.4 MAE worse than the no-heatmap baseline: even a noise-free density field degrades counting, so cue quality does not explain the failure. Removing the second image entirely drives the same checkpoint to 168.91, far worse than any adversarial content; the model has learned to depend on the second-image slot, and the gray condition (101.80), rather than image removal, is the appropriate in-distribution null.

\begin{table}[!htbp]
\caption{Adversarial heatmap-swap MAE on the dual-image SFT with valid P2PNet heatmaps. Spread is \textbf{23.39}, with the shuffled condition (a valid heatmap from a different image) the worst. The model is causally sensitive to the heatmap content but applies it counterproductively.}\label{tab:adversarial}
\centering
\begin{tabular}{lcc}
\toprule
Heatmap condition & MAE & $\Delta$ from correct \\
\midrule
Correct (P2PNet heatmap) & 96.26 & --- \\
Inverted (hot$\leftrightarrow$cold, same spatial pattern) & 97.85 & $+$1.6 \\
Gray (uniform, no spatial pattern) & 101.80 & $+$5.5 \\
Random noise & 103.12 & $+$6.9 \\
Shuffled (different valid heatmap) & \textbf{119.65} & $+$23.4 \\
\bottomrule
\end{tabular}
\end{table}

The visual channel is causally live but counterproductive: conditioning on a content-valid density signal modifies the prediction, yet the resulting integration degrades counting accuracy on every encoding tested. This is closer to active misuse than to passive inattention. The pattern is related but not identical to the multi-image underuse documented by MuirBench~\citep{wang2024muirbench}, which reports below-baseline performance of VLMs on tasks requiring information from secondary images. Where MuirBench characterises a quantitative attention deficit, the adversarial-swap protocol here surfaces a stronger qualitative failure: the secondary image is not merely under-attended but actively decoded in a way that worsens the count. We interpret this as a limitation of the late-fusion encoder pipeline at the spatial-field decoding level rather than at the multi-image attention level per se. The next subsection separates these explanations with two mechanism-targeted controls.

\subsection{Signal Abstraction, Not Channel Modality, Is the Binding Constraint}\label{sec:abstraction-controls}

The channel and abstraction-level explanations of the visual fusion failure are confounded in the experiments above. The visual channel carried a pre-integration density field; the text channel of CrowdCue-T carries an already-integrated integer and reaches overall MAE 72.04, with full results and statistical treatment in Section~\ref{sec:textinject-results}. We separate the two explanations with two mechanism-targeted controls.

\paragraph{CrowdCue-V (count-as-pixels).} The visual channel of the CrowdCue family (Section~\ref{sec:textinject}). The integer estimate $\hat{n}_P$ is rendered as printed digits on a plain white background ($448 \times 448$, monospaced) and supplied as the second image alongside the crowd photograph. The prompt does not mention the count. If the visual channel itself were the boundary, this configuration should fail in the same regime as the density-heatmap encodings of Section~\ref{sec:strategies}; if the boundary is the model's ability to decode dense spatial fields into integers, it should approach or exceed CrowdCue-T.

\paragraph{Per-region count injection.} A complementary control that varies signal richness rather than channel. The P2PNet point detector is rerun keeping point locations, and the points are bucketed into a $2 \times 2$ spatial grid. The four sub-counts (top-left, top-right, bottom-left, bottom-right) plus the total are injected into the text prompt as a structured estimate. The signal is intermediate in spatial richness between CrowdCue-T's single integer and the full density map of visual conditioning. A $4 \times 4$ variant extends the same construction to sixteen cells, listed row-major in the prompt, sampling a denser point on the same axis.

Table~\ref{tab:abstraction-controls} reports the result.

\begin{table}[!htbp]
\caption{Two mechanism controls separating channel from abstraction level, with the standalone specialist added for reference. CrowdCue-V (visual channel, integer symbol) yields the lowest MAE and RMSE of any configuration in the paper. The per-region text injection adds only marginal improvement over the single-integer text. The final column is overall RMSE across all 1{,}332 test images.}\label{tab:abstraction-controls}
\centering
\footnotesize
\begin{tabular}{@{}lcccccccc@{}}
\toprule
Configuration & Channel & Form & SHA-A & SHB & QNRF & NWPU & Overall & RMSE \\
\midrule
SFT-single (baseline) & --- & none & 70.17 & 15.47 & 119.19 & 102.55 & 81.64 & 352.66 \\
P2PNet specialist alone & --- & integer source & 61.26 & 21.93 & 121.88 & 107.41 & 84.45 & 221.93 \\
SFT-dual (valid heatmap) & visual & dense field & 71.16 & 13.93 & 153.61 & 119.13 & 96.26 & 443.13 \\
CrowdCue-T (Sec.~\ref{sec:textinject-results}) & text & integer & 66.08 & \textbf{11.74} & 113.44 & 84.65 & 72.04 & 249.83 \\
Per-region (text, $2 \times 2$) & text & 4 sub-counts + total & 71.74 & 12.80 & 115.20 & 78.36 & 71.19 & 191.47 \\
Grid (text, $4 \times 4$) & text & 16 sub-counts + total & 68.68 & 13.15 & 120.06 & 79.19 & 72.33 & 213.12 \\
\textbf{CrowdCue-V (this section)} & \textbf{visual} & \textbf{printed digits} & \textbf{65.21} & 13.25 & \textbf{106.96} & \textbf{63.33} & \textbf{62.65} & \textbf{163.22} \\
\bottomrule
\end{tabular}
\end{table}

CrowdCue-V reaches overall MAE 62.65, the lowest of any configuration we tested, ahead of CrowdCue-T by 9.4 MAE and the no-cue baseline by 19.0 MAE. It is also 21.8 MAE ahead of the specialist that supplies its cue (P2PNet alone reaches 84.45 on the same split; Table~\ref{tab:abstraction-controls}), so the VLM refines its tool rather than repackaging it, with the margin widest exactly where the specialist is weakest (NWPU 63.33 against 107.41, UCF-QNRF 106.96 against 121.88). The gain over the baseline is statistically firm: the mean per-image improvement is $+18.99$ MAE (paired bootstrap 95\% CI $[+7.71, +35.43]$; two-sided Wilcoxon $p = 1.5 \times 10^{-6}$), and overall RMSE drops from 352.66 to 163.22. The margin over CrowdCue-T is concentrated in the extreme-density tail: the mean difference is $+9.39$ MAE (CI $[+0.05, +21.76]$) while the median image is unchanged, and the aggregate ordering is reproducible, holding at all three seeds and under the STEERER specialist (Section~\ref{sec:ablations}). ShanghaiTech~B is the one dataset where the text variant stays ahead ($-1.51$ MAE, $p = 7 \times 10^{-4}$). CrowdCue-V wins on three of four datasets, with the largest single-dataset gain on NWPU (63.33 vs.\ 84.65, $-$25.2\%). The visual channel, when fed a pre-integrated integer rendered as printed digits, is therefore fully usable, and preferred in the aggregate where extreme-density images dominate the error.

The adversarial-override protocol confirms causal use. Replacing the rendered integer at evaluation time with 0, half, a random integer in $[0, 1000]$, or $10 \times$ the correct value drives MAE to 147.40, 167.12, 196.24, and 939.49 respectively, a 877-point spread across the five conditions, with format compliance at 100\% under every override for both variants. The spread is approximately $3 \times$ larger than the CrowdCue-T text-channel spread (304) and $37 \times$ larger than the dense-heatmap visual spread (23.4) reported in Table~\ref{tab:adversarial}. The raw ratios are indicative rather than exact, since the heatmap battery contains no perturbation as semantically large as a $10\times$ count override; the closest matched pair (shuffled heatmap at $+23.4$ against the random-integer override at $+61.0$ for CrowdCue-T and $+133.6$ for CrowdCue-V) preserves the ordering at smaller ratios. The visual channel anchors more strongly to printed integers than the text channel anchors to inline integers; the binding constraint in the dense-heatmap failures was decoding the field into a count, not reading the second image.

Two further results sharpen the comparison. First, anchoring strength is a two-sided property: the reliance that makes CrowdCue-V accurate under a correct cue also amplifies specialist failure, with the visual variant degrading further than the text variant under the zero and random overrides (147.40 and 196.24 against 96.90 and 133.08). A one-sentence prompt change recovers much of this margin: retraining CrowdCue-V with the verify-or-override license appended to its prompt caps the $10\times$ override at 581.42 against 939.49 without it, at no cost on correct cues. Second, the same wording-matched control separates prompt semantics from channel: the matched variant reaches MAE 64.21, within the seed spread of the original 62.65 (Section~\ref{sec:ablations}), so the license wording, the one textual difference between the two prompts, does not explain the visual variant's accuracy advantage.

A natural reading of CrowdCue-V is that the printed digits are transcribed by the vision encoder, in effect an optical character recognition (OCR) step, so the integer reaches the decoder as a symbol regardless of the channel that carried it. We agree with this reading, and it strengthens the conclusion rather than weakening it. The second-image position itself (the attention-disadvantaged slot documented by MuirBench) is not the obstacle, because a second image whose content survives transcription into a symbol is used more strongly than the same symbol placed inline in the prompt. CrowdCue-V therefore shows the visual channel works as a conduit for symbols. The claim is scoped to symbolic content: non-symbolic visual encodings of the same integer (e.g.\ a bar whose height encodes the count) are outside the tested design and outside the claim.

The per-region text injection sits between the two at MAE 71.19. Its overall gain over CrowdCue-T is small ($-$0.85), but concentrated on NWPU (78.36 vs.\ 84.65, $-$7.4\%), the extreme-density benchmark where a single integer most under-determines the answer. Spatial decomposition delivered through the text channel as quadrant counts helps slightly; the same spatial decomposition delivered as a density field through the visual channel hurts substantially (heatmap dual: 96.26). The asymmetry is in what the model can decode, not where it reads.

Taken together, the two controls invert the apparent contribution of channel modality. The signal that succeeds is the abstracted integer, through either channel. The signal that fails is the dense spatial field, in every visual encoding we tested; serialising it through the text channel does not rescue it either, since a $4 \times 4$ grid of sixteen sub-counts reaches 72.33, indistinguishable from the single integer's 72.04 (Table~\ref{tab:abstraction-controls}). Added spatial granularity through the text channel plateaus immediately, consistent with abstraction level rather than spatial richness being the operative variable. The MuirBench framing of multi-image attention as the failure axis does not fit our data: when the second image is a printed integer, the second-image route is the most effective conditioning route we found. We re-state the paper's central finding in Section~\ref{sec:discussion} accordingly.

\subsection{CrowdCue-T: Text-Channel Specialist Cueing}\label{sec:textinject-results}

We now report the text-channel variant of the CrowdCue family in full (Section~\ref{sec:textinject}). The pretrained P2PNet \citep{song2021p2pnet} produces an integer count $\hat{n}_{P}$ for each image; that number is injected verbatim into the SFT prompt as the specialist cue and the model is fine-tuned with standard SFT against ground-truth counts. We denote this configuration CrowdCue-T to distinguish it from the visual-channel variant CrowdCue-V evaluated in Section~\ref{sec:abstraction-controls}.

\paragraph{Specialist signal quality.} Before evaluating the VLM, we report the standalone accuracy of the injected signal on the combined test split. P2PNet (the official ShanghaiTech-A checkpoint applied uniformly across all four datasets) reaches SHA-A MAE 61.26 in-distribution and degrades on the out-of-distribution datasets: SHB MAE 21.93, NWPU 107.41, UCF-QNRF 121.88, overall 84.45. (Our SHA-A measurement differs from the 51.96 reported for the released checkpoint: our inference caps the longer image side at 2{,}048 pixels and rounds each dimension down to a multiple of 128 before the forward pass, with a confidence threshold of 0.5, whereas the official evaluation runs at full resolution. The checkpoint is trained on the SHA-A training images only, which are disjoint from every image we evaluate on.) The signal is therefore imperfect across all four benchmarks, which is the realistic operating point: the VLM has to learn when to trust the injected estimate and when to override it, rather than memorising a near-perfect prior.

\begin{figure}[htbp]
\centering
\includegraphics[width=\columnwidth]{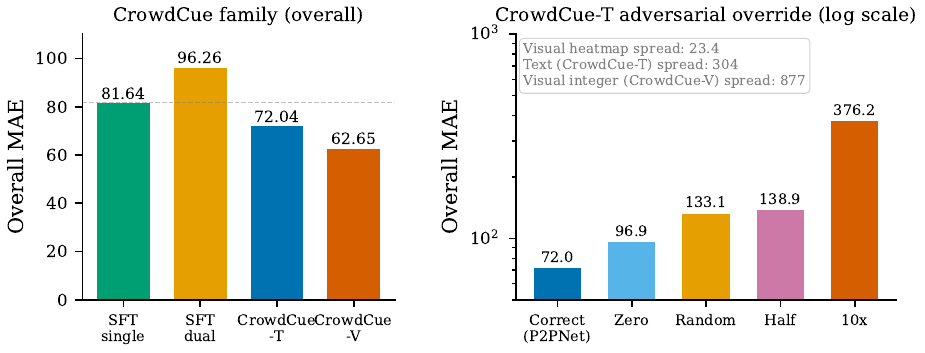}
\caption{Left: overall MAE for single-image, dual-image, and CrowdCue-T SFT models; the dashed line marks the single-image baseline. Right: MAE under the four adversarial text-channel overrides on a log scale. The 304-point spread under the $10 \times$ override contrasts with the 23.4-point spread across the five visual adversarial conditions on valid heatmaps (Section~\ref{sec:adversarial-results}).}
\label{fig:textinject-results}
\end{figure}

\paragraph{Main result.} Per-dataset numbers for every configuration appear in Table~\ref{tab:abstraction-controls}; Fig.~\ref{fig:textinject-results} plots the overall comparison. CrowdCue-T lowers overall MAE from 81.64 to 72.04 ($-$11.8\%) and from the visual dual-image baseline of 96.26 by 25.1\%. The headline mean improvement is $+9.60$ MAE per image (paired bootstrap, 5{,}000 resamples, 95\% CI $[-5.13, +29.14]$). The wide CI reflects heavy tails in per-image errors: a handful of extreme-density images contribute deltas in the thousands. A paired Wilcoxon signed-rank test, which is rank-based and insensitive to those tails, confirms the improvement ($n = 1{,}332$, statistic $= 400{,}481$; one-sided alternative that per-image absolute error under SFT-single exceeds that under CrowdCue-T; $p = 3.9 \times 10^{-5}$, or $7.9 \times 10^{-5}$ two-sided). Two robustness checks bound the magnitude estimate against the heavy tail. The median per-image improvement is $+1.00$ MAE (CI $[0, +1]$), so the typical image moves modestly in the CrowdCue-T direction. Winsorizing per-image deltas at the 1st and 99th percentiles constrains the mean to $+4.96$ MAE (CI $[-0.12, +10.22]$), demonstrating that the effect survives the heavy tail being bounded. The improvement is consistent in sign across every dataset, though individually significant only on SHA-B: SHA-A drops from 70.17 to 66.08 (CI $[-7.88, +16.25]$, includes zero), SHA-B from 15.47 to 11.74 (CI $[+2.25, +5.22]$, firmly positive), UCF-QNRF from 119.19 to 113.44 (CI $[-12.46, +23.65]$), and NWPU from 102.55 to 84.65 (CI $[-19.43, +68.21]$, the largest absolute drop with the widest CI because of extreme-density variance). The improvement is therefore consistent in direction, significant under the rank test, and robust to outliers; magnitude estimates are noisy because absolute errors concentrate in the extreme-density tail rather than because the effect itself is uncertain. Format compliance remains at 100\%.

\paragraph{Adversarial override.} Table~\ref{tab:textinject-adv} reports the same model evaluated with the four text-channel overrides defined in Section~\ref{sec:textinject}. A model that uses the injected number causally should shift its predictions in the direction of the override; a model that ignores it should remain near the correct-injection MAE.

\begin{table}[!htbp]
\caption{Adversarial overrides on CrowdCue-T, evaluated on the combined 1{,}332-image test split. $\Delta$ is the change in overall MAE from the correct injection.}\label{tab:textinject-adv}
\centering
\begin{tabular}{lcc}
\toprule
Inject override & MAE & $\Delta$ from correct \\
\midrule
Correct $\hat{n}_{P}$ & 72.04 & --- \\
$0$ & 96.90 & $+$24.9 \\
Random $\in \{0,\ldots,1000\}$ & 133.08 & $+$61.0 \\
Half $\hat{n}_{P}$ & 138.86 & $+$66.8 \\
$10 \cdot \hat{n}_{P}$ & 376.22 & $+$304.2 \\
\bottomrule
\end{tabular}
\end{table}

The overrides shift MAE by 25 to 304 points (Fig.~\ref{fig:textinject-results}), with the inflation pattern matching the magnitude of the perturbation: zero hurts least (the model can re-estimate from the image when given a clearly low anchor), random and half hurt more, and $10 \times$ creates pathological tail behaviour that the model cannot fully recover from. The contrast with the visual adversarial result on valid heatmaps (Table~\ref{tab:adversarial}, spread 23.4 points across five conditions) is the central methodological point of the paper: under identical training hyperparameters and the same specialist source, the integer is converted into accuracy gains while the dense field is read but misapplied. The variable that matters is the form of the signal, not the channel that carries it; Section~\ref{sec:abstraction-controls} established that the same integer succeeds through the visual channel as well.

The per-bucket breakdown (Table~\ref{tab:textinject-bucket}) clarifies where the channel matters most. In the sparse bucket the zero override leaves MAE essentially unchanged (16.85 vs 16.78); the model relies on the image when the injection looks implausible for the scene. In the very-dense bucket the model's reliance on the injected number jumps sharply: MAE 122.77 at the correct injection, 205.50 with zero ($+$82.7 points), 249.12 with $10\times$. In the extreme bucket the $10\times$ override produces catastrophic inflation (MAE 3{,}393) while zero leaves MAE only marginally worse (498 vs 380). At counts this large, the model has no way to recover from a wildly inflated anchor but can re-estimate from the image when the anchor is clearly too small. The pattern is consistent with the model learning to anchor on the specialist precisely where its own visual estimate is least reliable, and overriding it where it can resolve individuals directly. Two cells qualify the pattern. In the sparse bucket the random and half overrides in fact beat the correct cue (10.19 and 11.68 against 16.78), and the correct-cue sparse MAE is itself worse than the no-cue baseline (16.78 against 8.99 in Table~\ref{tab:bucket}). Where the model can resolve individuals, any anchor adds noise; the cue pays only where the visual estimate degrades.

\begin{table}[!htbp]
\caption{Per-density-bucket MAE for CrowdCue-T under correct and adversarial overrides. The override effect is concentrated in the very-dense and extreme buckets; the sparse bucket is nearly insensitive.}\label{tab:textinject-bucket}
\centering
\small
\begin{tabular}{@{}lcccccc@{}}
\toprule
Override & Sparse & Medium & Dense & Very dense & Extreme & Overall \\
\midrule
\textbf{Correct} & \textbf{16.78} & \textbf{13.71} & \textbf{50.67} & \textbf{122.77} & \textbf{379.70} & \textbf{72.04} \\
Zero & 16.85 & 19.30 & 56.62 & 205.50 & 498.37 & 96.90 \\
Half & 11.68 & 20.48 & 75.56 & 250.84 & 832.50 & 138.86 \\
Random & 10.19 & 20.73 & 90.55 & 206.67 & 783.17 & 133.08 \\
$10\times$ & 16.77 & 16.47 & 68.94 & 249.12 & 3393.13 & 376.22 \\
\bottomrule
\end{tabular}
\end{table}

\begin{figure}[htbp]
\centering
\includegraphics[width=0.8\columnwidth]{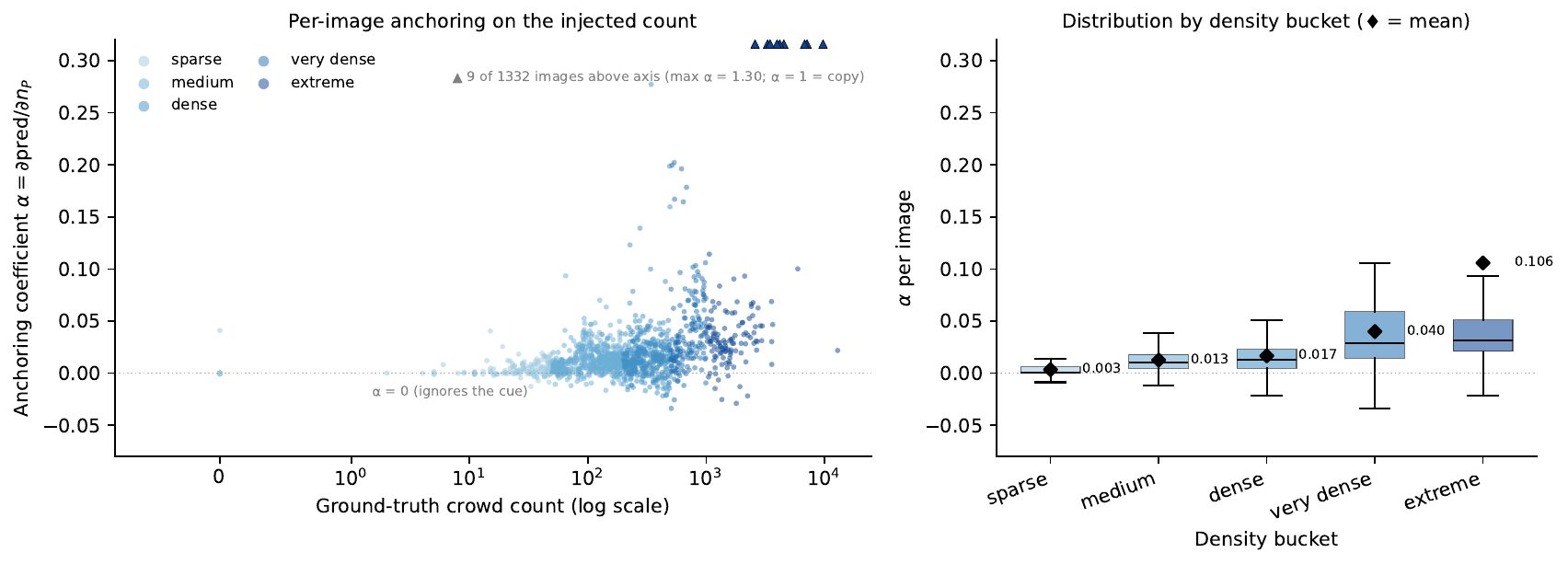}
\caption{Per-image anchoring coefficient $\alpha = \partial \hat{y} / \partial \hat{n}_{P}$ for CrowdCue-T, estimated by linear regression through the five injection values evaluated per image. Left: $\alpha$ against ground-truth count, coloured by density bucket (darker = denser). Right: per-bucket distribution, with diamonds marking bucket means. Mean $\alpha$ rises monotonically from near zero in sparse scenes (the model ignores the cue) to 0.106 in extreme scenes (the model anchors on it), an input-conditional trust policy learned from the gradient signal alone. The $\alpha$ axis is clipped at 0.32 for readability; nine of 1{,}332 images exceed the clip (maximum 1.30) and are drawn as triangles at the top edge.}
\label{fig:anchoring}
\end{figure}

\paragraph{Anchoring coefficient.} To quantify the trust policy directly, we estimate a per-image anchoring coefficient $\alpha = \partial \hat{y} / \partial \hat{n}_{P}$ by linear regression through the injection values evaluated for each image (correct, zero, half, random, and ten times the P2PNet count). Mean $\alpha$ rises monotonically from 0.003 in the sparse bucket to 0.106 in the extreme bucket, with an overall mean of 0.025, where 1 corresponds to copying the cue and 0 to ignoring it (Fig.~\ref{fig:anchoring}). The model is far from a copy machine at every density level, and the learned reliance rises exactly where its own visual estimate degrades. No branching instruction appears in the prompt or the training target; the input-conditional policy emerges from the gradient signal alone.

\subsection{Cross-Dataset Transfer and Training-Target Ablations}\label{sec:ablations}

We close with five analyses that bound the CrowdCue contribution: cross-dataset transfer, a training-target ablation, seed robustness, fixed-rule fusion baselines, and a specialist swap. The transfer experiment and training-target ablation were conducted on CrowdCue-T only; CrowdCue-V was identified as a mechanism control after these results landed and we leave the analogous V-variant transfer studies to future work (Section~\ref{sec:limitations}).

\paragraph{Cross-dataset transfer.} To test whether the cueing benefit survives a distribution shift, we retrain both SFT-single and CrowdCue-T on three datasets (SHA-A, SHB, UCF-QNRF; 1{,}901 training samples) with NWPU-Crowd held out, then evaluate on the full 1{,}332-image test set. Table~\ref{tab:xdataset} reports the result. The CrowdCue-T advantage transfers cleanly and amplifies under distribution shift. On the held-out NWPU specifically, CrowdCue-T lowers MAE from 133.99 to 106.18 (20.8\% improvement), substantially exceeding the 17.5\% gain CrowdCue-T shows on full-data NWPU (Table~\ref{tab:abstraction-controls}). The overall improvement under cross-dataset training (25.5\%; 109.99 $\to$ 81.97) is more than double the full-data improvement (11.8\%), consistent with the cueing benefit being most pronounced exactly where the in-distribution training mix is weakest.

\begin{table}[!htbp]
\caption{Cross-dataset transfer: SFT on SHA + SHB + QNRF only (1{,}901 samples), evaluated on the full test set. NWPU is held out from training. The CrowdCue-T advantage transfers: it leads on the held-out NWPU and on all three in-distribution datasets.}\label{tab:xdataset}
\centering
\begin{tabular}{@{}lccccc@{}}
\toprule
Config & SHA-A & SHB & QNRF & \textbf{NWPU (held out)} & Overall \\
\midrule
SFT single (excl-NWPU) & 80.03 & 22.70 & 172.96 & 133.99 & 109.99 \\
CrowdCue-T (excl-NWPU) & \textbf{70.84} & \textbf{13.78} & \textbf{116.34} & \textbf{106.18} & \textbf{81.97} \\
\bottomrule
\end{tabular}
\end{table}

\paragraph{Training target: flat scene descriptor versus verify-and-adjust.} We initially designed CrowdCue-T with an explicit verify-and-adjust SFT target: the assistant \texttt{<think>} block referenced the injected number directly and branched on its relative error against the ground truth. The version reported throughout Sections~\ref{sec:textinject-results} and above uses a strictly simpler target, the single-image bucket descriptor with no reference to the injected estimate, and the simpler target wins on three of four datasets and overall (Table~\ref{tab:training-target}; verify-and-adjust stays ahead on UCF-QNRF). The injected number in the user prompt is the load-bearing element; the verify-and-adjust scaffolding in the assistant target was apparently over-prescribing the reasoning trace, leaving the model less anchored on the injection. The simpler variant also responds more strongly to the mild overrides (a 24.9-point shift under the zero override against 10.7 for verify-and-adjust), consistent with tighter anchoring on the cue; under the extreme $10\times$ override the verify-and-adjust variant inflates further (461.9 against 304.2).

\begin{table}[!htbp]
\caption{Training-target ablation. Both variants use the same image input and the same text-injection prompt; only the assistant SFT target differs. The flat target reaches lower MAE on every dataset.}\label{tab:training-target}
\centering
\begin{tabular}{@{}lccccc@{}}
\toprule
Assistant target & SHA-A & SHB & QNRF & NWPU & Overall \\
\midrule
Verify-and-adjust template & 74.15 & 13.46 & 110.08 & 91.23 & 75.17 \\
Flat scene descriptor (CrowdCue-T) & \textbf{66.08} & \textbf{11.74} & 113.44 & \textbf{84.65} & \textbf{72.04} \\
\bottomrule
\end{tabular}
\end{table}

\paragraph{Seed robustness.} We retrain the three headline configurations with two additional seeds (43, 44). Overall MAE across seeds 42/43/44: SFT-single 81.64/88.61/81.75 (mean 84.00), CrowdCue-T 72.04/67.73/70.72 (mean 70.16), CrowdCue-V 62.65/65.21/62.12 (mean 63.33). The per-configuration ranges (6.97, 4.31 and 3.09 MAE) are small against the cue effect, and the ordering CrowdCue-V $<$ CrowdCue-T $<$ SFT-single holds at every seed. The seed means separate more cleanly than the seed-42 point estimates (84.00 against 70.16 against 63.33). Headline numbers elsewhere in the paper are the seed-42 runs.

\paragraph{Fusion baselines.} The cueing framing invites a simpler explanation: perhaps any combination of the two underlying estimators would do as well. We compare CrowdCue against fixed-rule combinations of its own two inputs, computed from the stored per-image predictions of the SFT-single model and the P2PNet specialist (Table~\ref{tab:fusion}). CrowdCue-T (72.04) is ahead of every realistic fixed rule: the naive mean of the two predictors (74.13), a linear blend fitted by five-fold cross-validation (87.80), and a per-bucket gate that picks the better predictor per P2PNet-predicted density bucket (75.42). An oracle gate that selects per ground-truth bucket reaches 70.00 but requires the true density at test time; CrowdCue-V (62.65) is ahead of even this bound. The SFT recipe therefore implements an input-conditional fusion that fixed rules do not match. Two cells bound the claim honestly. On NWPU, CrowdCue-V (63.33) sits far below both of its components (102.55 and 107.41), which no convex combination of the two can reach. On in-distribution SHA-A, the standalone cue (61.26) is ahead of both CrowdCue variants (66.08 and 65.21); fusion pays where the specialist is weak and costs a little where the specialist is strongest.

\begin{table}[!htbp]
\caption{CrowdCue against fixed-rule fusion baselines computed from stored per-image predictions of the same two estimators (SFT-single and P2PNet). The gate baselines pick the better predictor per density bucket, using either the P2PNet-predicted bucket (realistic) or the ground-truth bucket (oracle). CrowdCue-T is ahead of every realistic rule; CrowdCue-V is ahead of the oracle bound.}\label{tab:fusion}
\centering
\begin{tabular}{@{}lcc@{}}
\toprule
Predictor & MAE & RMSE \\
\midrule
P2PNet alone & 84.45 & 221.93 \\
SFT-single alone & 81.64 & 352.66 \\
Mean of the two & 74.13 & 256.37 \\
Fitted linear blend (5-fold CV) & 87.80 & 259.59 \\
Per-bucket gate (P2PNet-predicted bucket) & 75.42 & 217.03 \\
Per-bucket gate (ground-truth bucket, oracle) & 70.00 & 195.19 \\
\midrule
CrowdCue-T & \textbf{72.04} & 249.83 \\
CrowdCue-V & \textbf{62.65} & \textbf{163.22} \\
\bottomrule
\end{tabular}
\end{table}

\paragraph{Specialist swap.} To test sensitivity to the choice of specialist, we replace the P2PNet count with the count from a per-dataset STEERER \citep{han2023steerer} checkpoint at evaluation time, with no retraining (1{,}331 of 1{,}332 test images; STEERER's standalone MAE on this split is 43.57). CrowdCue-V improves from 62.65 to 60.89 and CrowdCue-T from 72.04 to 70.21, and the visual variant stays ahead of the text variant under the new specialist. Under swap-only evaluation the fused model does not recover the stronger specialist's standalone accuracy. We attribute this to training: both variants were fine-tuned on P2PNet-quality cues and learned a reliance calibrated to that error level. Retraining with STEERER cues confirms it. CrowdCue-T retrained on STEERER training-split counts reaches overall MAE 45.47 and CrowdCue-V reaches 42.50 (1{,}331 images), the visual variant again ahead, and the retrained CrowdCue-V edges past the specialist alone (43.57) by 1.1 MAE, a margin we read as parity rather than superiority. Two qualifiers apply: the STEERER cues come from per-dataset checkpoints, a stronger and dataset-aware signal than the single SHA-A P2PNet used everywhere else, and these are single-seed runs. Read at that tier, the family scales with its specialist rather than being capped by it, and the bundled language output comes at no measured MAE cost.

\subsection{Per-Density Analysis}\label{sec:perdensity}

Table~\ref{tab:bucket} breaks down SFT performance by crowd density.

\begin{table}[!htbp]
\caption{MAE by density bucket for SFT single-image model. Error scales superlinearly with crowd count.}\label{tab:bucket}
\centering
\begin{tabular}{@{}lrrr@{}}
\toprule
Density Bucket & Count Range & $n$ (test) & MAE \\
\midrule
Sparse & 0--50 & 171 & 8.99 \\
Medium & 50--200 & 498 & 16.20 \\
Dense & 200--500 & 378 & 52.89 \\
Very dense & 500--1{,}000 & 160 & 144.20 \\
Extreme & $>$1{,}000 & 125 & 448.56 \\
\bottomrule
\end{tabular}
\end{table}

The error profile is sharply superlinear: moving from sparse to medium roughly doubles the MAE (9 $\to$ 16), but moving from dense to extreme increases it ninefold (53 $\to$ 449). At the extreme end, with crowds exceeding 1{,}000 people, the model systematically undercounts because it lacks the spatial aggregation mechanism that density regression networks use to integrate over continuous fields. Below 200 people, however, the VLM achieves errors comparable to or better than many purpose-built counting methods, suggesting that VLMs can perform reasonable counting in scenes where individual features remain at least partially resolvable at $512 \times 512$ resolution.

\subsection{Zero-Count Accuracy}

Among 35 NWPU-Crowd test images with zero people, the SFT single-image model correctly predicts zero for 33 images (94.3\%). The two false positives produce mean predictions of 30.4 (single) and 44.0 (dual). The two specialists measured in this paper illustrate the contrast on the same 35 images: P2PNet emits exactly zero on 3 of 35 (MAE 209.0, worst case 2{,}334 in an empty scene) and the per-dataset STEERER on 4 of 35 (MAE 10.7). Density regression integrates over textured backgrounds and rarely lands on exactly zero. The VLM's semantic understanding allows it to recognize that a scene contains no people, bypassing the noise-integration failure mode inherent to continuous density estimation.

\subsection{Inference Timing}

On a single H100 GPU with greedy decoding, the SFT single-image model processes each test image in 1.00\,s. The dual-image configuration requires 1.26\,s (+26\%), reflecting the additional visual tokens. By comparison, CSRNet processes an image in approximately 50\,ms, a 20$\times$ speed advantage. VLM-based counting is not competitive for real-time applications but is practical for offline analysis tasks where language-based output is valued.


\subsection{Contextual Comparison with Specialised Methods}\label{sec:context}

Table~\ref{tab:context} situates our results against four families: the 2025 SOTA specialists (ZIP, DSGC-Net, RCCFormer), the 2024 diffusion baseline CrowdDiff, the CLIP-based hybrid CLIP-EBC, and the historical CSRNet. Two observations matter.

\begin{table}[!htbp]
\caption{VLM counting against the 2025 specialist front. Numbers are MAE; --- denotes a result the original paper did not report. The last column flags methods that emit free-form reasoning alongside the count. ZIP-P matches our 7B VLM on SHA-A with 0.81M parameters; CrowdCue-V closes most of the baseline-to-ZIP-B gap on NWPU and roughly a quarter of it on the other datasets. Our NWPU column uses the validation split (test labels are not public), while published NWPU results are test-server numbers where available, so NWPU cells are not split-matched across rows. The P2PNet row reports our measurement with the SHA-A checkpoint applied to all four datasets (Section~\ref{sec:textinject-results}).}\label{tab:context}
\centering
\small
\begin{tabular}{@{}lrccccl@{}}
\toprule
Method & Params & SHA-A & SHA-B & QNRF & NWPU & Free-form output \\
\midrule
ZIP-B \citep{ma2025zip} & 105.6M & \textbf{47.8} & \textbf{5.5} & \textbf{69.4} & \textbf{60.1} & no \\
DSGC-Net \citep{wu2025dsgcnet} & $\sim$30M & 48.9 & 5.9 & --- & --- & no \\
RCCFormer \citep{chen2025rccformer} & --- & 48.3 & 6.6 & 77.6 & 74.3 & no \\
CrowdDiff \citep{ranasinghe2024crowddiff} & --- & 47.4 & --- & --- & --- & no \\
CLIP-EBC \citep{clipEBC2025} & --- & 55.0 & 6.3 & --- & --- & partial \\
ZIP-P \citep{ma2025zip} & \textbf{0.81M} & 71.1 & --- & --- & --- & no \\
CSRNet \citep{li2018csrnet} & $\sim$16M & 68.2 & 10.6 & --- & --- & no \\
P2PNet \citep{song2021p2pnet} (our specialist) & --- & 61.26 & 21.93 & 121.88 & 107.41 & no \\
\midrule
Ours (SFT, single) & 7B & 70.17 & 15.47 & 119.19 & 102.55 & yes \\
Ours (CrowdCue-T) & 7B & 66.08 & 11.74 & 113.44 & 84.65 & yes \\
Ours (CrowdCue-V) & 7B & 65.21 & 13.25 & 106.96 & 63.33 & yes \\
\bottomrule
\end{tabular}
\end{table}

First, the parameter axis. ZIP spans more than two orders of magnitude in backbone size; its smallest variant (ZIP-P, 0.81M parameters) reaches SHA-A MAE 71.1, essentially matching our 7B Qwen2.5-VL fine-tune at 70.17. The takeaway is not that the VLM is wasteful; it is that MAE-per-parameter is the wrong axis on which to ask whether a generalist model has earned its place. A general-purpose VLM is where you go when the deployment needs free-form reasoning, instruction-following, or semantic gating (e.g., emitting zero for an empty scene) alongside the count.

Second, the accuracy gap. On SHA-A the 2025 specialist front sits at MAE $\sim$48; our SFT-single at 70.17 trails by roughly 22 MAE points, the same gap that separated CSRNet (68.2) from today's frontier, i.e., roughly seven years of specialist progress. CrowdCue narrows the gap further: CrowdCue-V at 65.21 sits about 17 MAE behind ZIP-B on SHA-A, and reaches 63.33 on NWPU against ZIP-B's 60.1, though the two NWPU numbers are not split-matched (ours is the validation split, ZIP-B's is presumably the test server). One 7B backbone is shared across all four datasets. We do not claim parity, and the underlying-specialist comparison is asymmetric: ZIP-P trained on 300 SHA-A images, while CrowdCue trains on 5{,}010 multi-dataset images and consumes a SHA-A-trained P2PNet's hints. The contribution this paper makes is methodological: diagnosing the dense-field decoding bottleneck and showing that an abstracted-integer cue, delivered through either channel, recovers a substantial fraction of what visual conditioning failed to deliver (Sections~\ref{sec:adversarial-results}, \ref{sec:abstraction-controls}, \ref{sec:textinject}). Any future VLM-based counter, or any VLM-specialist hybrid, will need to confront this same constraint.

\subsection{Qualitative Examples}

\begin{figure}[htbp]
\includegraphics[width=1\columnwidth]{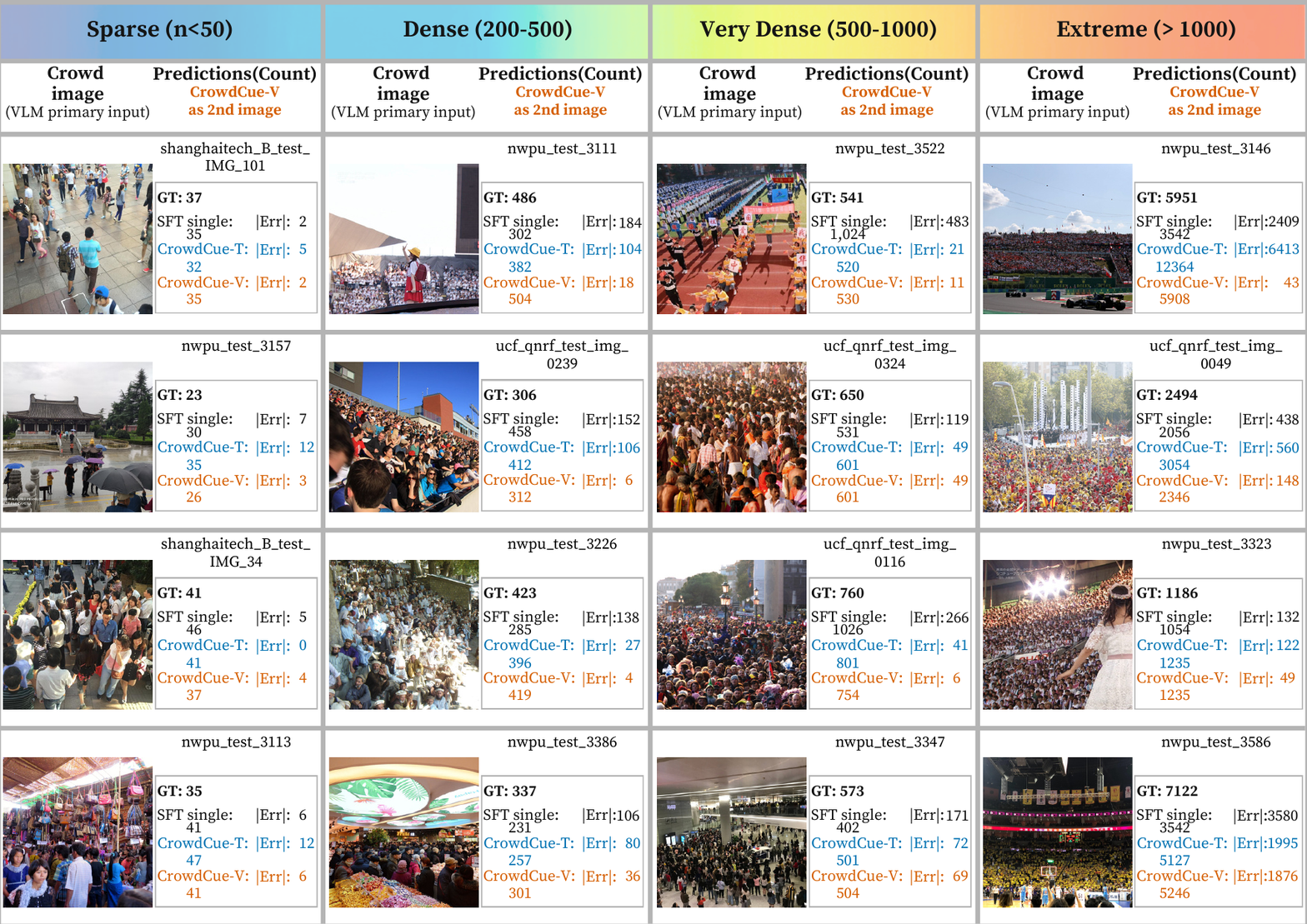}
\caption{Qualitative comparison of SFT-single, CrowdCue-T, and CrowdCue-V predictions across four density levels, with ground truth and per-model absolute error for each example.}
\label{fig:qualitative}
\end{figure}

Fig.~\ref{fig:qualitative} presents representative predictions from SFT-single, CrowdCue-T, and CrowdCue-V across four density levels.

\section{Discussion}\label{sec:discussion}

\subsection{Why Dense-Field Conditioning Fails and Integer Conditioning Works}

The central empirical finding of this paper is that signal abstraction, not channel modality, governs whether the model converts auxiliary specialist information into accuracy gains in our setting. The configurations that make the case (Tables~\ref{tab:visual-encodings-valid}, \ref{tab:abstraction-controls} and~\ref{tab:textinject-adv}) share the same backbone, training recipe, dataset and specialist source, and differ only in the form and route of the auxiliary signal: the integer succeeds through either channel, while the density field fails in every visual encoding tested. Three forces explain why the integer succeeds where the field does not.

\textbf{Architectural alignment.} Qwen2.5-VL is a late-fusion VLM: each image passes independently through the ViT and produces its own token sequence, and cross-image relationships only emerge in the LLM decoder's self-attention. The decoder is optimised for sequential reasoning over a single text stream, not for geometric alignment of two visual token sequences. A second image entering this architecture has no architectural prior linking its tokens to corresponding regions in the first image; an integer entering as text drops directly into the sequence the decoder was trained to consume. MuirBench \citep{wang2024muirbench} documents the same weakness across VLM families, including the positional bias summarised in Section~\ref{sec:related}. Our finding is consistent with a structural property of late-fusion pipelines rather than a Qwen-specific accident, though replication across architectures is required before it can be claimed as one (Section~\ref{sec:limitations}).

\textbf{Optimisation pressure.} SFT minimises next-token prediction loss. When the crowd image alone is sufficient to reach MAE 81.64, the optimiser has no incentive to learn the cross-image alignment needed to extract per-region information from a second image. The heatmap tokens are dead weight that the model learns to route around. The text-channel route changes the loss landscape: an integer in the prompt is directly comparable to the integer in the target, and the model can either copy it (when accurate) or revise it (when off), both of which lower the loss measurably. The architectural barrier disappears and the optimisation signal is concentrated on a single token relationship. This is consistent with \citet{hayat2025attention}, who show that VLMs preferentially rely on the modality that most directly minimises training loss.

\textbf{Where the channel matters most.} The per-bucket adversarial breakdown shows the model leans on the text-channel anchor exactly where its independent visual estimate is weakest. In the sparse and medium buckets, the overrides shift MAE by at most 7 points; in the very-dense bucket they move it by between 83 and 128 points, and in the extreme bucket by hundreds. The injected count substitutes for the spatial aggregation that VLMs cannot perform reliably above $\sim$500 individuals, while leaving the model's intrinsic capability untouched where it is already accurate. This is the asymmetry that makes the text channel a useful conditioning route rather than a wholesale replacement: the model learns when to defer and when to override.

\textbf{Alternative explanations.} Three deserve statement. First, the heatmap encoding bounds what any model could recover: each density map is normalised to $[0,1]$ per image before colormapping (Section~\ref{sec:heatmap}), which discards the multiplicative scale, so the count is not recoverable from the heatmap alone and the signal can act only as a spatial prior. The swap protocol shows the model reads that spatial structure and misapplies it, so the information ceiling does not explain the failure by itself, but it caps what dense-field conditioning of this form could have delivered. Second, jet-colormapped scalar fields are data visualisations, a category far from the ViT's pretraining distribution, and part of the failure may sit in the encoder embedding rather than in cross-image attention. Third, all three dense encodings share one rendering family (fixed-bandwidth Gaussian splat, jet colormap), so the negative result is established for that family rather than for dense fields in general.

\subsection{What VLM Counting Gets Right}

The conditioning study should not obscure the genuine capabilities the VLM brings.

\textbf{Sub-200 counting is competitive.} With MAE 8.99 for sparse scenes and 16.20 for medium density (single-image SFT), the VLM is competitive in the sparse regime, where individual heads are at least partially resolvable at $512 \times 512$; at street-scene densities the specialist front remains stronger (SHB 5.5--6.6 against our 11.74), though CrowdCue-V retrained with per-dataset STEERER cues reaches SHB 5.55 (Section~\ref{sec:ablations}). Text-injection trades some accuracy on the sparse bucket (16.78) for substantial gains in the heavy tail; for deployments dominated by low-density scenes, the single-image checkpoint is the right choice.

\textbf{Zero-count detection exploits semantic gating.} The 94.3\% accuracy on the 35 NWPU zero-count images reflects scene-level understanding: the VLM recognises an empty scene and emits zero, sidestepping the false-positive integration problem that density regressors face on textured backgrounds. On the same 35 images, our two measured specialists emit exact zeros on 8.6\% (P2PNet) and 11.4\% (STEERER) of cases against the VLM's 94.3\% (Section~\ref{sec:experiments}); ZIP's zero-inflated likelihood targets this failure mode from the regression side but we did not measure it.

\textbf{SFT fully resolves format compliance.} After SFT, structured output in the \texttt{<think>}/\texttt{<answer>} format is produced 100\% of the time, compared to 0.8--13.4\% in the zero-shot setting. The format itself must be learnt; once learnt, it persists across the visual conditioning interventions reported here and the text-injection variant.

\subsection{Limitations}\label{sec:limitations}

Five limitations bound the conclusions drawn here.

\textbf{Single specialist.} The injected signal in Section~\ref{sec:textinject-results} is a P2PNet count from the official SHA-A checkpoint, applied uniformly across all four datasets. Its standalone MAE ranges from 61.26 (in-distribution SHA-A) to 121.88 (UCF-QNRF). A stronger specialist (e.g., ZIP \citep{ma2025zip} at 47.8 on SHA-A) would likely tighten the text-injection result further, particularly on the heavy tail; an in-distribution specialist per dataset would tighten it more. We deliberately use a single, modestly-accurate specialist to test whether the channel itself, rather than perfect signal quality, drives the effect.

\textbf{Resolution constraint.} All experiments operate at $512 \times 512$ pixels due to the interaction between visual-token count and training context length. Naive upscaling to $768$ or $1024$ did not improve MAE in preliminary trials; an end-to-end resolution-aware reconfiguration of training (longer context, fewer steps per epoch, possibly a different LoRA target set) might.

\textbf{Single VLM architecture for the headline results.} Our headline results use Qwen2.5-VL-7B. A preliminary cross-architecture check with the same SFT recipe supports partial generality: on InternVL3-8B, the text-channel cue lowers overall MAE from 93.08 (single image) to 59.96, with 100\% format compliance and causal cue use (a zero override degrades MAE to 105.82). LLaVA-OneVision-7B fails the format prerequisite (compliance below 78\% single-image and below 28\% with the cue), so its accuracy is not comparable. The cueing benefit replicates on one additional late-fusion architecture while the format prerequisite does not replicate on another; broader replication, and the transfer to early-fusion VLMs, remain open.

\textbf{SFT-only training.} Both visual conditioning and text injection are trained via SFT. GRPO with a reward referencing the injected count, or DPO on verify-vs-override pairs, might amplify the cueing benefit further. An initial GRPO attempt with the fuzzy reward of \citet{wang2025crowdvlm} layered on CrowdCue-T dropped format compliance below 65\%, so format-preserving constraints (e.g.\ a stronger format-reward weight, a KL penalty against the SFT reference, or DPO on paired format-valid completions) appear necessary before reinforcement learning can stack on the family safely.

\textbf{Seed replication is partial.} The three headline configurations are replicated across three seeds (Section~\ref{sec:ablations}), with overall-MAE ranges of 3.1--7.0 and the CrowdCue-V $<$ CrowdCue-T $<$ SFT-single ordering preserved at every seed. All other configurations are single runs, so orderings that rest on gaps below a few MAE points (e.g.\ the per-region control against CrowdCue-T, or the flat against the verify-and-adjust target) should be read as indistinguishable.

\subsection{Toward 3D-Aware CrowdCue Extensions}

The results suggest that the most useful auxiliary signal for a late-fusion VLM is not a dense spatial field but an already-integrated symbolic cue that the model can easily anchor, verify, and adjust.
A natural future direction is therefore to enrich the symbolic cue rather than to add another raw visual modality.
In particular, 3D-aware crowd density estimation could provide geometry-informed summaries that go beyond a single global count.
Digital-twin environments, calibrated multi-camera systems, stereo/depth sensing, or LiDAR-assisted reconstruction could estimate crowd occupancy in physical coordinates and then serialize that information as textual or numerical cues, such as counts per depth band, floor level, stadium tier, entrance/exit region, or occlusion zone.
This would preserve the abstraction advantage observed in CrowdCue-T and CrowdCue-V while adding spatial structure that a single integer cannot capture.

This direction is most relevant to the high-count tail, where 2D projections compress depth, perspective, and occlusion onto a single image plane. The superlinear error growth above 1{,}000 people (Table~\ref{tab:bucket}) suggests that the model's visual estimate becomes increasingly unreliable precisely when geometric priors are most useful.
However, our results also caution against passing raw 3D fields, depth maps, or volumetric heatmaps directly to the VLM as visual inputs; such representations may reproduce the same dense-field decoding failure observed with 2D heatmaps. The more promising question is whether 3D estimators can produce structured CrowdCue-style summaries that help the VLM reason over extreme-density scenes while remaining within a symbolic conditioning regime.

\subsection{Integration with Sensor-Based Crowd Monitoring}


The CrowdCue family extends to non-visual sensor signals without architectural change. WiFi probe counts, Bluetooth Low Energy beacon density, pressure-sensitive floor mats, and CO\textsubscript{2} concentration estimates already exist in deployed monitoring stacks and serialise to short text strings (e.g., ``WiFi probe density: 347; BLE devices observed: 192''). Our single P2PNet integer is the simplest case of this pattern; a multi-sensor extension would inject a feature vector (i.e., counts, densities, and confidence intervals where available) and let the model weight each reading against the visual scene. The argument that motivated the text channel (i.e., late-fusion VLMs underuse parallel visual streams but consume sequential text natively) applies a fortiori to modalities that carry no visual representation at all. We have not run this multi-sensor experiment; whether the cross-modality weighting emerges from SFT alone or requires explicit fusion supervision in the training target is the obvious next test.


\subsection{Ethical Considerations}\label{sec:ethics}


Four deployment concerns deserve explicit acknowledgement.

The first is privacy. CrowdCue estimates aggregate counts rather than identifying individuals, but the underlying VLM retains the face-recognition and person-re-identification capabilities of its pretraining; a deployed system inherits those latent capabilities even when the task asks only for a count. Architectural safeguards (i.e., freezing or pruning identification-relevant attention heads, or distilling the counting behaviour into a smaller dedicated decoder) are required to commit a deployment to counting-only functionality.

The second is demographic bias. Density regressors and VLMs alike can exhibit differential accuracy across crowd compositions and cultural contexts (e.g., varying clothing, head coverings, body spacing norms). Our evaluation does not disaggregate errors along any of these axes; the existing crowd counting benchmarks do not carry the demographic metadata that would let us. A bias audit on a demographically annotated test set is a precondition we have not met, not one we have cleared.

The third is dual-use exposure. Crowd counting at scale supports benign public-safety management and equally supports the surveillance applications most ML venues now ask authors to address. We do not believe a single paper can resolve that tension. We do believe public-space deployments should be gated on transparent policy disclosure, human oversight, and the compliance framework that applies in the specific jurisdiction (e.g., GDPR, EU AI Act).

The fourth is specific to the CrowdCue family: the injected cue is an attack surface. The override results (Section~\ref{sec:textinject-results}) show that a corrupted upstream specialist shifts the reported count by design, so a deployment must treat the specialist's integrity as part of the system's trusted base.


\section{Conclusion}\label{sec:conclusion}

To our knowledge, this work delivers the first fine-tuned generative-VLM crowd counting evaluation spanning ShanghaiTech A and B, UCF-QNRF, and NWPU-Crowd. SFT is necessary: zero-shot format compliance sits below 14\%, and fine-tuning lifts it to 100\% at overall MAE 81.64. Dense-density-heatmap conditioning fails in every visual encoding tested, and the adversarial-swap protocol shows the model reads the heatmap but misapplies it. The same specialist signal succeeds as an already-integrated integer: CrowdCue-T reaches MAE 72.04 through the text channel and CrowdCue-V reaches \textbf{62.65} as a printed digit image, the lowest result in the paper and 21.8 MAE ahead of the specialist that supplies the cue. The binding constraint in our setting is the model's ability to decode dense spatial fields into counts, not the channel through which the signal arrives. The improvement transfers across distributions (a 20.8\% gain on held-out NWPU), survives a specialist swap, and is robust to random seed and to the training-target choice (Section~\ref{sec:ablations}).

The methodological contribution generalises beyond crowd counting. The adversarial-override protocol, paired across visual and textual channels with the same specialist source, isolates the channel as the causal variable in VLM conditioning experiments. Any future VLM-specialist hybrid (for counting, for detection-quality refinement, for document understanding) can be tested with the same procedure.

The accuracy frontier is not where this work makes its claim. ZIP-B \citep{ma2025zip} reaches SHA-A MAE 47.8 with 105M parameters; ZIP-P reaches MAE 71.1 with 0.81M parameters, essentially matching our 7B fine-tune on raw accuracy. The value proposition of the generative VLM is the bundled capability stack (structured reasoning traces, instruction-following, 94.3\% zero-count semantic gating, free-form language output) that no density regressor offers natively. Specialist-cue conditioning narrows the accuracy gap enough that the bundled stack becomes a defensible choice for deployments where the count is one of several requested outputs.

Several directions follow. Three-dimensional density estimators from digital-twin reconstructions could supply richer per-region integer summaries than the single integer studied here, particularly for the extreme-density regime where 2D projections lose depth. Non-visual sensor modalities (WiFi probes, BLE beacons, pressure mats) serialise naturally to integers and inherit the abstraction-level advantage demonstrated by both CrowdCue-T and CrowdCue-V; hybrid visual-plus-sensor injection through the CrowdCue family is the obvious next experiment.

Pre-deployment auditing for VLM-based counters extends the override protocol of Section~\ref{sec:textinject-results} from numerical perturbations to demographic-shift inputs (varying crowd composition, attire, and lighting). Differential accuracy across groups is undocumented at present and is the kind of property any public-space deployment review will demand, alongside privacy safeguards against the underlying VLM's latent face-recognition capabilities. Two further directions deserve note. Early-fusion VLM architectures, in which visual tokens from multiple images interact within the vision encoder, would test whether the dense-field decoding failure documented here is fundamental to the late-fusion family or a property of current implementations; the success of CrowdCue-V already establishes that the visual channel itself is not the obstacle. Reinforcement learning with a reward referencing the injected count is a natural extension, subject to the format-compliance constraint discussed in Section~\ref{sec:limitations}.

\FloatBarrier

\bibliographystyle{plainnat}
\bibliography{references}

\end{document}